\documentclass[sigconf]{acmart}

\AtBeginDocument{%
  }

\usepackage{amsmath,amsfonts,amsthm}
 
\usepackage{amssymb}
\usepackage{mathtools,nccmath}
\usepackage{mathrsfs}

\usepackage{subfigure}
\usepackage{caption}
\usepackage{graphicx}
\usepackage{textcomp}
\usepackage{booktabs}
\usepackage{multirow}
\usepackage{url}
\usepackage{numprint}
\usepackage{ragged2e}
\usepackage{tabularx}
\usepackage{natbib} 
\usepackage[inline]{enumitem}
\usepackage{tikz}
\usepackage{xcolor,color,soul,bm}
\usepackage{xspace}
\usepackage{lipsum}

\usepackage{amssymb}
\usepackage{pifont}
\newcommand{\cmark}{\ding{51}}%
\newcommand{\xmark}{\ding{55}}%
\newcommand{\METHOD}{\texttt{CroSSL}}

\newcommand{\mdl}{\mathrm{X}}
\newcommand{\glabel}{\mathrm{Y}}

\newcommand{\MI}{\mathtt{I}}
\newcommand{\Ent}{\mathtt{H}}
\newcommand{\given}{~|~}

\usepackage{xspace}
\newcommand{\latinphrase}[1]{\textit{#1}}

\newcommand{\ie}{\latinphrase{i.e.,}\xspace}
\newcommand{\eg}{\latinphrase{e.g.,}\xspace}

\newcommand{\revisioncolor}{black}

\newcommand{\revision}[1]{\textcolor{\revisioncolor}{#1}}

\copyrightyear{2024}
\acmYear{2024}
\setcopyright{rightsretained}
\acmConference[WSDM '24]{Proceedings of the 17th ACM International Conference on Web Search and Data Mining}{March 4--8, 2024}{Merida, Mexico}
\acmBooktitle{Proceedings of the 17th ACM International Conference on Web Search and Data Mining (WSDM '24), March 4--8, 2024, Merida, Mexico}\acmDOI{10.1145/3616855.3635795}
\acmISBN{979-8-4007-0371-3/24/03}

\begin{document}

\title{\METHOD{}: Cross-modal Self-Supervised Learning \\ for Time-series through Latent Masking}

\author{Shohreh Deldari}\authornote{Work has been done during the author's internship at Nokia Bell Labs.}
\affiliation{%
  \institution{University of New South Wales}
  \city{Sydney}
  \country{Australia}}
\email{s.deldari@unsw.edu.au}

\author{Dimitris Spathis}
\affiliation{%
  \institution{Nokia Bell Labs}
  \city{Cambridge}
  \country{UK}}
\email{dimitrios.spathis@nokia-bell-labs.com}

\author{Mohammad Malekzadeh}
\affiliation{%
  \institution{Nokia Bell Labs}
  \city{Cambridge}
  \country{UK}}
\email{mohammad.malekzadeh@nokia-bell-labs.com}

\author{Fahim Kawsar}
\affiliation{%
  \institution{Nokia Bell Labs}
  \city{Cambridge}
  \country{UK}}
\email{fahim.kawsar@nokia-bell-labs.com}

\author{Flora D. Salim}
\affiliation{%
  \institution{University of New South Wales}
  \streetaddress{1 Th{\o}rv{\"a}ld Circle}
  \city{Sydney}
  \country{Australia}}
\email{flora.salim@unsw.edu.au}

\author{Akhil Mathur}
\affiliation{%
  \institution{Nokia Bell Labs}
  \city{Cambridge}
  \country{UK}}
\email{akhilmathurs@gmail.com}

\renewcommand{\shortauthors}{Shohreh Deldari et al.}

\begin{abstract}

Limited availability of labeled data for machine learning on multimodal time-series extensively hampers progress in the field. Self-supervised learning (SSL) is a promising approach to learning data representations without relying on labels. However, existing SSL methods require expensive computations of negative pairs and are typically designed for single modalities, which limits their versatility. We introduce \METHOD{} (Cross-modal SSL), which puts forward two novel concepts: {\em masking intermediate embeddings} produced by {\em modality-specific encoders}, and their aggregation into a {\em global embedding} through a {\em cross-modal aggregator} that can be fed to downstream classifiers. 
 \METHOD{} allows for handling {\em missing modalities} and end-to-end {\em cross-modal} learning without requiring prior data preprocessing for handling missing inputs or negative-pair sampling for contrastive learning. We evaluate our method on a wide range of data, including motion sensors such as accelerometers or gyroscopes and biosignals (heart rate, electroencephalograms, electromyograms, electrooculograms, and electrodermal) to investigate the impact of masking ratios and masking strategies for various data types and the robustness of the learned representations to missing data. 
 Overall, \METHOD{} outperforms previous SSL and supervised benchmarks using minimal labeled data, and also sheds light on how latent masking can improve cross-modal learning. Our code is open-sourced at \url{https://github.com/dr-bell/CroSSL}

\end{abstract}



\maketitle

\section{Introduction}

Sensory signals captured through {\em multiple modalities} of smart devices, such as accelerometers, gyroscopes, and electroencephalography (EEG), facilitate various applications; ranging from human activity recognition~(HAR)~\cite{tang2021selfhar} to sleep tracking via brain-activity monitoring ~\cite{kemp2000analysis}. Many of such emerging applications are based on machine learning~(ML) techniques and particularly deep neural networks~(DNNs).
 However, the reliance on \textit{labeled} data for training DNNs has hindered their ability to scale effectively~\cite{yuan2022self}. Due to the high cost and time demands of gathering, annotating, and managing extensive labeled datasets, self-supervised learning~(SSL), which learns from unlabeled data, has been investigated~\cite{saeed2019multi}: by defining an artificial task, known as a \textit{pretext} task, where the supervisory signal is automatically generated from unlabelled data, enabling the training of an encoder model to learn a latent representation of the input data~\cite{yuan2022self}. SSL has demonstrated practicality in HAR~\cite{tang2021selfhar}, by leveraging large amounts of unlabeled data and fine-tuning on downstream tasks with limited labeled data.

However, existing SSL methods are mostly designed for \textit{unimodal}~\cite{haresamudram2020masked, saeed2019multi, tang2021selfhar, xue2021exploring} data and struggle to handle \textit{multimodal} data~\cite{haresamudram2020masked, saeed2019multi, tang2021selfhar, xue2021exploring}. Particularly in numerous health monitoring and physiological applications, data is often acquired from \textit{heterogeneous} sensors of various modalities with different characteristics (e.g., sampling rates and resolution). Current SSL methods show inadequate performance when it comes to aggregating and compressing a time-window of various sensors into a coherent {\em global embedding} that can properly serve a downstream task~\cite{deldari2022cocoa}. Creating embeddings that incorporate multiple modalities becomes even more challenging due to the dynamic nature of real-world situations in which the granularity of sensor data and availability of modalities can differ from one user to another or from time to time.

By collecting and analyzing data from diverse sources, we can inform and improve our understanding of human behavior and physiology. For example, in the case of elderly health monitoring, a 360-degree health-monitoring system that combines images captured by smart glasses, audio signals captured by smart earbuds, and sensor time series captured by smartwatches can provide a wealth of information about the user's physical and cognitive state~\cite{olmedo2022remote}. Here, multi-modal SSL can distill the combined data into a unified inference engine that can facilitate several downstream tasks, such as the early detection of falls, the prediction of cognitive decline, and the monitoring of sleep quality, and physical activity levels. Moreover, multi-modal SSL enables the discovery of complex interconnections and correlations among multiple data sources, which can provide a more comprehensive understanding of human sensing. For instance, as humans, we naturally learn to identify objects in our surroundings through observations made using our multiple senses. In the absence of input from some senses, we can still recognize the object through our remaining senses, demonstrating the power of multi-modal integration. 

Notably, to design multimodal SSL in real-world settings we face two major challenges. (1) \emph{Heterogeneous Sensors}: different sensors require different preprocessing due to their data characteristics and different sampling rates. Direct integration of heterogeneous sensor data leads to inconsistencies in the global embeddings and unsatisfactory performance~\cite{yan2021deep,shwartz2023compress}. 
(2) \emph{Missing Sensors}: multimodal SSL must be robust to missing modalities. During SSL training, a model might learn representations that rely on correlations between different modalities; however, there is no guarantee that all the modalities are available at inference time. 
\revision{These two challenges are partially addressed by a prior work COCOA~\cite{deldari2022cocoa} via the concept of modality-specific encoders and a customized loss function to align latent embeddings across different modalities. However, COCOA does not perform aggregation on the modality-specific embeddings and does not consider the challenge of missing modalities. Thus, as we show, COCOA struggles to produce useful global embeddings for downstream tasks when some modalities are missing.}

To fill this gap, we propose \METHOD \space  (Cross-modal Self-Supervised Learning) by incorporating two novel concepts of (i)  masking intermediate embeddings produced by modality-specific encoders, and (ii) creating global embeddings for potential downstream classifiers by a cross-modal aggregator; thus offering three major contributions: 
(1) \METHOD{} is a novel SSL technique that incorporates latent masking in order to learn robust cross-modal embeddings of data, without requiring any additional data preprocessing on the input level. 
(2) \revision{We evaluate \METHOD{} on multi-modal sensing datasets including HAR, health-monitoring, and stress and affect detection tasks with various types of sensors such as accelerometer, gyroscope, heart rate, electroencephalogram (ECG), electromyograms (EMG), electrooculograms (EEG), and electrodermal (EDA) sensors. 
We conduct systematic experiments to investigate the impact of different masking ratios and masking strategies (\eg random or spatial) for different data types and show to which extent learned embeddings can be robust to missing modality experiments. Furthermore, we investigate using minimal labeled data (ranging from 1\% of data) in learning informative embeddings for applications with limited access to annotated data.}
(3) Our work sheds light on the impact of masking on learning from multimodal signals.

\section{Related Work}
\label{sec:related}

There are two common \textit{pre-training} approaches: (1) utilizing models pre-trained on labeled data from a different task, or (2) leveraging unlabeled data from the same task~\cite{yuan2022self}. The former is not applicable to healthcare applications due to limited labeled data and heterogeneity of sensors, but the latter has shown effectiveness in learning general and transferable features~\cite{oord2018representation, chen2020simple}.
\textit{Supervised} learning works explore fusion strategies and adapted DNNs for HAR~\cite{radu2018multimodal}. Unsupervised or self-supervised learning approaches are also applied to sensor signals, focusing on HAR, including multi-tasking, contrastive learning, and predictive coding~\cite{saeed2019multi, tang2021selfhar, tonekaboni2021unsupervised, haresamudram2021contrastive}. None of these works are specifically designed for \textit{multimodal} learning, where we need representations that capture both sensor-specific temporal dependencies and global spatial dependencies across sensors. ColloSSL~\cite{jain2022collossl} and COCOA~\cite{deldari2022cocoa} address some aspects of multimodal learning but have limitations such as non-trivial negative pair mining and the inability to handle missing modalities. Missing modalities are challenging as they result in incomplete or biased data, and learning representations that generalize across modalities is difficult due to different distributions and feature spaces.

Incorporating various modalities is even more complex due to the absence of shared information and the challenges associated with alignment and integration. Masking techniques have been proposed mostly for vision and text data \cite{he2022masked, }. MultiMAE~\cite{bachmann2022multimae} uses input-level masked autoencoders which however are designed for image data and lack support for multimodal time series. While masking is straightforward in the input data space, it is not commonplace in the latent space. As seen in recent speech models such as TERA and Wav2Vec~\cite{liu2021tera, baevski2020wav2vec}, a temporal mask is first randomly applied in the latent space, where ~50\% of the projected latent feature vectors are dropped. While our approach is conceptually similar to these, we improve this idea within a multimodal architecture that learns from different views of two masks. This is crucial, as not all modalities may be available during training or inference. Addressing these challenges, we incorporate latent masking and modality-specific encoders to learn joint representations of multimodal healthcare data, enabling efficient integration and robustness to missing modalities. Table~\ref{tab_related_work} summarizes the conceptual differences between \METHOD{} and related prior work in the context of HAR and health monitoring applications.

\begin{table}[]
\caption{Existing SSL methods in ubiquitous sensing.
}\label{tab_related_work}
\resizebox{0.48\textwidth}{!}{%
\begin{tabular}{lccccc}
Method & SSL method & Loss $f(.)$& Cross-modal & { Missing data}\\ \hline
TPN (2019)~\cite{saeed2019multi}  & pretext (transformation)  & BCE & \xmark&\xmark\\
Haresamudram~(2020)~\cite{haresamudram2020masked}    &  pretext (masking) & MSE & \xmark&\xmark\\
Sense\&Learn (2021) \cite{saeed2021sense}  & pretext (transformation) & CE + MSE & \xmark&\xmark\\
SelfHAR~(2021)~\cite{tang2021selfhar}    & self-distillation & CCE & \xmark&\xmark\\
TNC (2021)~\cite{tonekaboni2021unsupervised}   & contrastive learning & Triplet loss & \xmark&\xmark\\
CPC (HAR) (2021)~\cite{haresamudram2021contrastive}    & contrastive learning & infoNCE & \xmark&\xmark\\
Yuan et al (2022)~\cite{yuan2022self}  & pretext (transformation) & BCE & \xmark&\xmark \\
ColloSSL (2022)~\cite{jain2022collossl}   & contrastive learning & infoNCE &  \cmark&\xmark\\
COCOA (2022)~\cite{deldari2022cocoa}  & contrastive learning & based on InfoNCE&  \cmark& \xmark\\
\hline
\textbf{\METHOD ~(our work)}  & pretext \& regularization & VICReg &  \cmark & \cmark\\ \hline 
\end{tabular}%
}
\end{table}

\section{Proposed Method: \METHOD{}}

\begin{figure*}
    \centering
    \includegraphics[width=\linewidth]{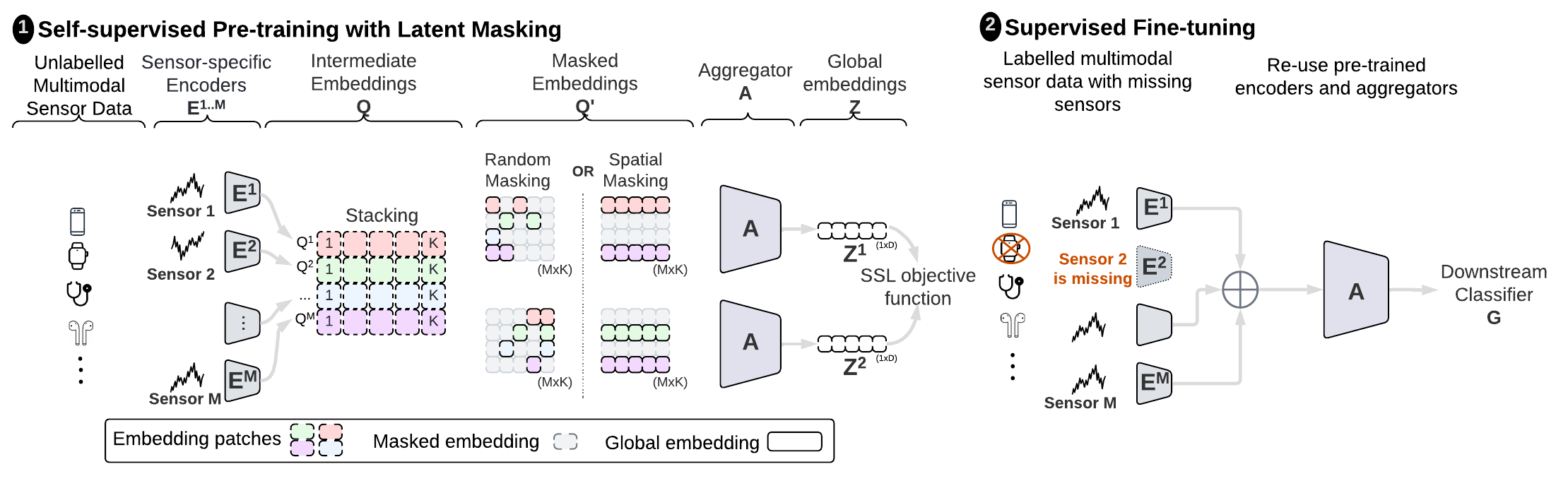}
    \caption{The overview of the proposed architecture. 
    }
    \label{fig_method}
\end{figure*}

\subsection{Problem Formulation}
{\bf Notation.} {Let $ X_{i}=\{X_{i}^{1},\cdots,X_{i}^{M}\}, i \in \{1,\cdots, N\}$ denote time-windows of multivariate time-series where $N$ is the number of training samples in each batch, and $M$ is the number of {\em data modalities} (\ie sensors/sources) captured by smart devices or biosensors.  
}
Let $\textbf{E}^m$ denote an {\em encoder} that maps data $X_i^{m}$ of modality $m$ into {\em intermediate embeddings}, $Q^m$ of size $K$, and $\textbf{A}$ denote the {\em aggregator} that maps intermediate embeddings of all encoders into a unified $D$-dimensional {\em global embedding} $Z_{i}=\{z_{i,1}, \cdots, z_{i,D}\}$ for each $X_{i}$. Let $\textbf{G}$ denote a downstream classifier that takes $Z_i$ as the input and makes the final prediction. For the purpose of brevity, we drop the index $i$ unless it is required.

\noindent
{\bf Goal.} A naive solution combines data from all modalities into a single encoder, but it does not work when the data types ($X^{m}$) and sizes ($T$) vary. An alternative is to use separate encoders  ($E^m$) and classifiers ($G^m$) for each modality and aggregate predictions. This approach enhances robustness but is expensive and does not leverage information from multiple modalities. Our goal is to  propose a solution that facilitates the usage of multiple heterogeneous modalities without requiring multiple classifiers.

\noindent
{\bf Solution Overview.} We build upon the SSL paradigm with the aim of capturing both intra-modality (i.e., temporal) and inter-modality (i.e., spatial)  dependencies. The objective is to build a {\em unified global embedding} from available data sources. Ideally, we want to compress information captured by each modality so that the aggregated information can be efficiently and accurately used for any downstream tasks (e.g., classification).  \revision{To this end, we assume that data captured by different modalities at the same time could be interpreted as natural transformations of each other, where all modalities are sensing the same phenomena.} Thus, such multimodal data, even in the absence of any annotation, can be leveraged by SSL methods \revision{due to the phenomena that are shared among all of them}.   In our setting, in contrast to the traditional SSL approaches, the supervisory signal does not come from only ``self'' \cite{haresamudram2021contrastive} or ``other'' \cite{deldari2022cocoa} sources of data but from the aggregation of (any subset of) sources.  \revision{In sum, our hypothesis is built upon the fact that (i) although individual sensors provide complementary views of the same event, they may not record the same information, or (ii) some sensors might not be relevant in some situations. However, (iii) the aggregated information offered by the global embeddings can represent the current state of the shared event. }

\noindent
{\bf \METHOD: Cross-modal Self-Supervised Learning}. Figure~\ref{fig_method} illustrates the overall architecture which consists of self-supervised pre-training (left panel) and fine-tuning (right panel).
Mainly, our goal in \METHOD{} is to utilize unlabelled and asynchronous data collected from different modalities and devices, and learn  {\em modality-specific encoders} $\textbf{E}^m$, $m\in\{1,2,\dots,M\}$, followed by a {\em cross-modal aggregator} $\textbf{A}$. Each $\textbf{E}^m$ is trained to generate an informative intermediate embedding for sensor $m$, \revision{in a way that the aggregation of all (or part of) intermediate embeddings can represent the state of the shared phenomena}. To this end, the {\em aggregator} is trained to learn both {\em spatial} and {\em temporal} dependencies across input data sources. Secondly, we aim to use all pre-trained encoders and the aggregator to obtain descriptive embeddings for a small amount of labeled data and subsequently train a classifier $\textbf{G}$  which maps these latent joint embeddings to the corresponding class labels. Our self-supervised pre-training is depicted in Figure~\ref{fig_method}.

Finally, in Step 2, we fine-tune the pre-trained encoder and aggregator module along with the downstream task. We use the pre-trained encoders $\textbf{E}_m$ and the aggregator $\textbf{A}$ and a labeled dataset of the given task to train the classifier $\textbf{G}$ (or any other downstream task) in a supervised fashion. 
The evaluations (Section \ref{sec:evaluation}) revealed that this latent embedding masking recipe for pre-training offers greater robustness compared to other baselines in dealing with lower-quality data that contains missing information during either or both fine-tuning or inference steps.

\noindent
{\bf Self-supervised Objective function}.
\revision{The dominant SSL techniques used in computer vision and natural language processing are based on either contrastive-based or reconstruction-based objectives. In contrastive learning models, the objective is to minimize the distance between positive pairs (\eg such as two augmented views of the same image) while repelling the representations of negative pairs apart (\eg augmented views of different images). 
The primary vulnerabilities of contrastive learning models lie in the quality of the positive and negative pairs utilized during training. In applications such as human activity recognition or emotion recognition, with limited variety in the number of classes and tasks compared to vision tasks, the probability of comparing samples with fake negatives during contrastive training is high \cite{deldari2021time}. Fake negative pairs are essentially positive pairs that are incorrectly labeled as negative pairs due to non-ideal sampling techniques, resulting in the model mistakenly attempting to push them apart. Although several negative mining techniques are proposed to remove fake negatives or avoid the biases caused by their existence in the final contrastive objective function \cite{NEURIPS2020_63c3ddcc,robinson2021hardcontrastive}, they are shown to be not effective in wearable sensor data \cite{deldari2021time}.}

Recently, there has been an emergence of regularization-based SSL techniques that do not require negative sampling, such as BYOL \cite{niizumi2021byol}, Barlow-Twins \cite{zbontar2021barlow}, and the most recent one VICReg which has been shown to be more effective in representation learning \cite{bardes2022vicreg}. 
As we utilize a variant of the VICReg (Variance-Invariance-Covariance Regularization) loss function, we briefly introduce this loss function and how we integrated it into our multi-sensor setup. For sample $X_i$, we generate two global embeddings $Z_i = (Z_{i}^{1},Z_{i}^{2})$, denoted as $ Z_i^{j}= \{z_{i,1}^{j},\cdots,z_{i,D}^{j}\}$ where $i$ is the number of corresponding original sample in the batch and $D$ is the dimension of embeddings coming out of the two branches of {\em aggregator}. 
The optimization function is based on three parts including:
\begin{itemize}[leftmargin=*]
    \item \revision{\textit{Invariance}: Minimize the distance (or maximize similarity) between the global embeddings of sample $Z_i$:}

    \begin{equation} 
        \label{eq:invariance_term}
        \revision{
        s(Z^1,Z^2) = \frac{1}{N}\sum_{i=1}^{N} \frac{1}{D}\sum_{j=1}^{D}{\| z_{i,j}^{1}-z_{i,j}^{2}\|_2^2}}
    \end{equation}

    \item \revision{\textit{Variance}: Maintain variance of each variable of embeddings denoted by $z_{i,j}^{1|2}$, $j=1\cdots d$ above a threshold. The variance regularization term is defined as:}
    \begin{equation}
        \label{eq:variance_term}
        \revision{ v(Z^{\{1|2\}}) = \frac{1}{D} \sum_{j=1}^{D}{ max\big(0, \gamma- \delta (z_{i..N,j}^{1|2},\epsilon)\big)}
        }
    \end{equation} 
    \revision{where $\gamma>0$ is a constant threshold for the regularized standard deviation term $\delta (z,\epsilon) = \sqrt{Var(z+\epsilon)}$ and $\epsilon$ is a scalar to avoid numerical instabilities. Here, $z_{i..N,j}$ shows that the standard deviation is computed for the $j$-th variable of the global embedding across the batch of size $N$. This term prevents collapsed representation by encouraging the variance across each variable of the global embedding in the batch to be equal to $\gamma$. }

    \item \revision{\textit{Covariance}: Minimize correlation between variables of the same embedding by minimizing the covariance as below:}

    \begin{equation}
        \label{eq:covariance_term}
        \revision{ c(Z) = \frac{1}{D} \sum_{j=1, j\neq j'}^{D}{[C(Z)]_{j,j'}^{2}}
        }
    \end{equation} 
    \revision{where $C(Z)$ is the covariance matrix of $Z$, and the covariance regularization
term  $c(Z)$ is the sum of the squared off-diagonal coefficients of the covariance matrix with a factor of $1/D$. This forces the off-diagonal elements to zero to decorrelate the embedding variables and maximize the distribution across variables.}
\end{itemize}

\begin{figure}
    \centering
    \includegraphics[width=0.5\textwidth]{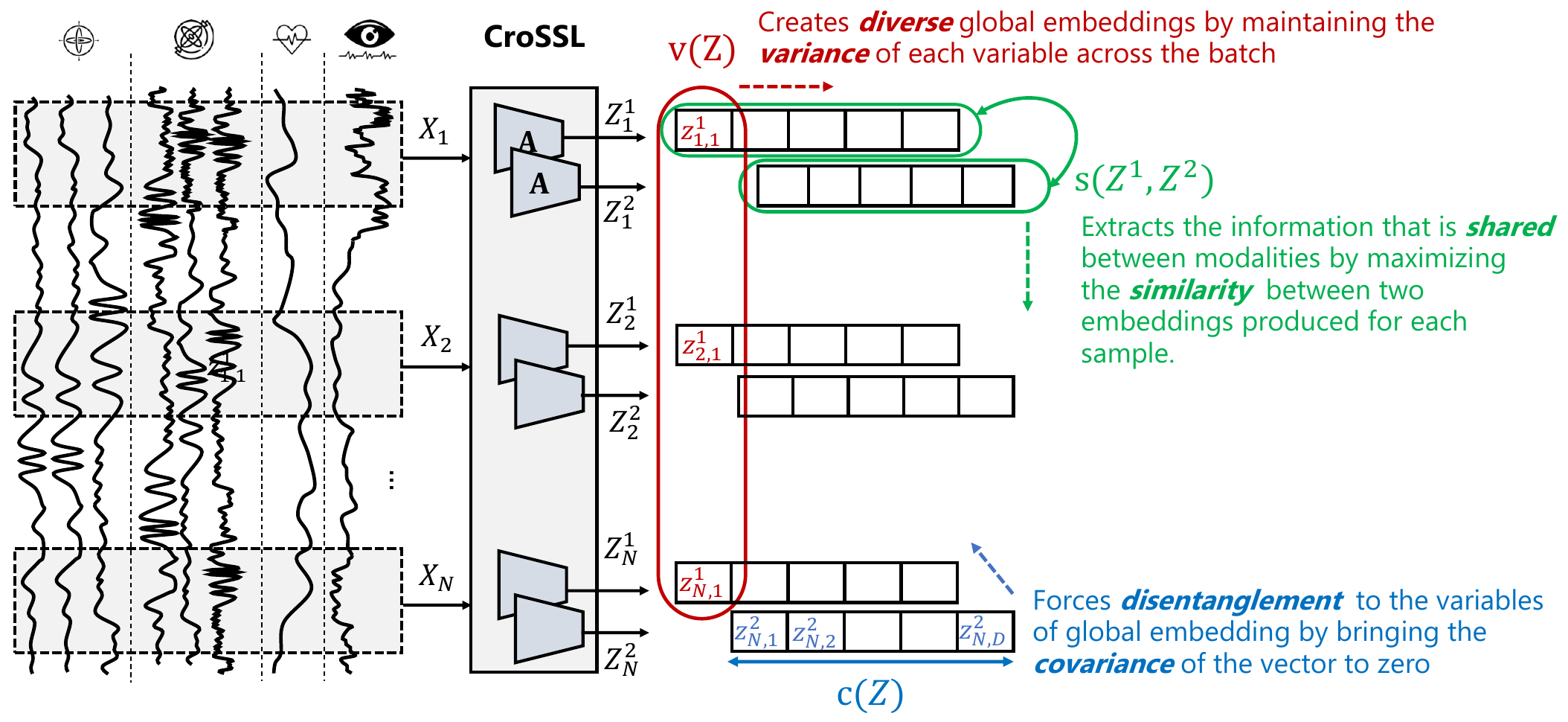}
    \caption{Overview of the regularization-based objective function within the \METHOD{} architecture.}
    \label{fig:loss_fn}
\end{figure}

\revision{Figure \ref{fig:loss_fn} illustrates our objective function. The overall loss function is a weighted average of the above terms \footnote{\revision{All the constant values are set as suggested by \cite{bardes2022vicreg}. Hyper-parameters $\gamma, \mu$, and $\nu$ have been set as 10, 10, and 100 through grid search hyper-parameter tuning.}}:}

\begin{equation*}
        \label{eq:loss_fn}
        \revision{ \mathcal{L}(Z^1, Z^2) = \lambda s(Z^1,Z^2)+ \mu [v(Z^1)+v(Z^2)] + \nu [c(Z^1)+c(Z^2)]
        }
    \end{equation*}

\subsection{Theoretical Motivations} 
\label{sec:theory}
In \METHOD{}, we consider that different modalities sense common phenomena and share a common high-level semantic, despite the fact that they might not necessarily capture the same amount of information. 
But, not all modalities are universally available, relevant, or helpful in real-world applications, and their usefulness varies depending on the situation and the specific task. 

\METHOD{} assumes that there are two types of information available in each intermediate embedding $Q^m$: (1) {\em cross-modal information} that is about the shared phenomena captured by all modalities, and (2) {\em modality-specific information} that are not necessarily relevant to the shared phenomena. \METHOD{} aims to produce global embedding $Z$ that carry cross-modal information between each sensor data $X^i$ and at least one other sensor $X^j$. Translating the above intuition into an information-theoretic formulation, we have
$$
\MI(\mdl^i;\mdl^j) \geq \epsilon > 0 \text{\quad for any~~} i \in [M]  \text{~~and a~~} j \in [M]/i,
$$
where $\MI(\cdot;\cdot)$ denote the mutual information between two random variables \revision{(\eg two randomly-sampled time-windows of sensor data)}. In many downstream tasks, the data $X^i$ is seen to be generated through a latent-variable generative process~\cite{tian2019contrastive, federici2020learning}. Basically, $\mdl^i = \mathcal{G}^i(\glabel)$, where $\glabel$ is the latent variable: a common source of variation that affects the data generated by all modalities \revision{(\eg the user's {\em activity} or {\em wellness} is the unknown-but-common source of variation for all modalities that are sensing the user)}. Making inferences about such a common object or subject is usually the ultimate target of the downstream task when processing the global embedding.

With this assumption, we are interested in learning an aggregator $A$ to generate a global embedding $Z$ that satisfies:
\begin{equation}\label{eq_mi_z_1}
\MI(Z;Q^i \given Q^j) = 0, \text{\quad for any~~} i \in [M]  \text{~~and a~~} j \in [M]/i.
\end{equation}
This means that we aim to generate $Z$ such that it ideally captures only the information shared between both $Q^i$ and $Q^j$, and not information specific to only one of them. \revision{In other words, bringing conditional mutual information
$\MI(Z;Q^i \given Q^j)$ to zero implies that} given $Q^j$, we can recover the information that is needed to generate $Z$ without needing $Q^i$. To understand this, we note that the chain rule for mutual information allows the following:
\begin{equation}
\MI(Z; Q^i, Q^j) = \MI(Z; Q^j) +
\MI(Z; Q^i \given Q^j).
\end{equation}
Thus we can rewrite~\eqref{eq_mi_z_1} as
\begin{equation*}
\begin{split}
\MI(Z; Q^i \given Q^j)
= \MI(Z; Q^i, Q^j) - \MI(Z; Q^j)   \\
= \MI(Z; Q^i) - \MI(Z; Q^j)
+ \MI(Z; Q^j \given Q^i) 
\end{split}.
\end{equation*}
Therefore, our assumption can be also interpreted as generating a global embedding $Z$ that satisfies the following:
\begin{equation}
\MI(Z; Q^i) = \MI(Z; Q^j).
\end{equation}
In this way, the generated global embedding $Z$ will ideally stay the same even if we miss one of our modalities. 

We emphasize that in \METHOD{}, we do not want to preserve all the information present in all modalities. In other words, we do not want trivial solutions where $Z$ is just a compressed version of data provided by all the modalities, \revision{and to prevent this, we provide the aggregator with two distinct masked versions of the intermediate embeddings.} Our approach is closer to the {\em Information Bottleneck} principle \cite{michael2018on}, which sets an objective to build $Z$ such that it includes only a small amount of information carried by each $X^m$, which is only relevant to $\glabel$. However, the challenge, in practice, is that most of the encoders $\textbf{E}$ are deterministic and can preserve an infinite amount of information. Notice that $\MI(Q^i,X^i) = \Ent(Q^i) - \Ent(Q^i \given X)$, and for a deterministic $\textbf{E}$ we have $\Ent(Q^i \given X^i)=0$. 

\revision{To retain cross-modal information in $Z$ and disregard modality-specific information, we employ {\em randomization} in generating $Z$  during training. This randomization is introduced through masking strategies, where certain modalities are masked, forcing the aggregator to rely on the remaining modalities for extracting cross-modal information. This ensures that the most informative modalities are used to generate the global embedding $Z$ for downstream tasks. Note that without randomization, the aggregator would produce identical embeddings by copying information from Q to Z, without distinguishing cross-modal from modality-specific information.}

\section{Evaluation}
\label{sec:evaluation}

\begin{table}
\caption{Characteristics of the datasets used in our evaluation.}
\label{tab:dataset-table}
\centering
\resizebox{0.48\textwidth}{!}{
\begin{tabularx}{\textwidth}{lcccX}
\textbf{Dataset} & \textbf{Sensors}   &\textbf{Subjects} & \textbf{Size} & \textbf{Classes \{Number\} }\\ \midrule
PAMAP2   & \begin{tabular}[c]{@{}c@{}}Acceleration\\  Gyroscope \\ (arm, chest, ankle) \\ \revision{Heart rate} \end{tabular}         & 9 &11K   &     \begin{tabular}[l]{@{}l@{}}   \textit{lying, sitting, standing, walking, running, cycling, nordic walking, watching TV,}\\  \textit{computer work, car driving, ascending stairs, descending stairs, vacuum cleaning, } \\ \textit{ironing, folding laundry, playing soccer, rope-jumping} \{12\}   \end{tabular}  \\ \hline
\begin{tabular}[c]{@{}c@{}}SLEEPEDF\\ \end{tabular}          & \begin{tabular}[c]{@{}c@{}}2xEEG, EOG, EMG\end{tabular}            &38 & 55K  &  \textit{Awake, Rapid Eye Movement, N$1$, N$2$-N$3$}, and \textit{N}$4$    \{5\}         \\ \hline
\begin{tabular}[c]{@{}c@{}}\revision{WESAD}\\ \end{tabular}          & \begin{tabular}[c]{@{}c@{}}\revision{ECG, EMG, EDA}\end{tabular}            &\revision{15} & \revision{21K}  &  \revision{\textit{baseline, stress, amusement}, and \textit{meditation}  \{4\}    }     \\ 
\bottomrule
\end{tabularx}
}
\end{table}

\subsection{Datasets}
To evaluate our approach, we study \revision{three} different datasets with various types of sensors and applications, including PAMAP2 \cite{reiss2012pamap2} a human activity recognition dataset based on motion and heart sensors mounted on different parts of the body; PhysioNet Sleep-EDF \cite{kemp2000analysis,goldberger2000physiobank} 
dataset for sleep stage detection based on biosensor data\revision{, and WESAD \cite{wesad}, a dataset for stress and affect recognition using different types of biosensors}. Table \ref{tab:dataset-table} provides details on the number of sensors, subjects, classes, and data size. 

\subsection{Experiments Setup}

Encoders can be chosen appropriately depending on the type of sensors. In this work, we use three layers of a 1D convolution network as sensor-specific encoders and fully connected layers for the aggregator module. However, the proposed framework is encoder-agnostic, which means the backbone encoder can be replaced with any other encoder model appropriately chosen according to the type of each sensor.
Following the evaluation framework in \cite{tian2019contrastive,jain2022collossl}, a linear classifier was used to evaluate the quality of pre-trained model and extracted embeddings. The pipeline for the downstream task training is depicted in Figure \ref{fig_method}-Step 2.

Our training setup is implemented in Tensorflow 2.0. We used the TF HParams API
for hyperparameter tuning and arrived at the following training hyper-parameters: \{ssl learning rate: 1e-4, classification learning-rate: 1e-3,  $\tau = 0.05$\}. We also explored a range of coefficients for the VICReg objective function, varying between \{1,10,100\} for each of variance, invariance and covariance coefficients; after hyper-parameter tuning, we noticed that the combination of \{variance coefficients: 10, invariance coefficients: 10, covariance coefficients: 100\} provides the most stable results.

We use Adam optimizer and early stopping to stop the training after five epochs without progress. All models are trained for 100 epochs during self-supervised learning and 50 epochs for during the supervised classification. When fine-tuning the model, we must prevent the encoders from forgetting their learned parameters and being overwritten by the classifier loss function. To achieve this, we freeze the encoders and the aggregator for the first 20 epochs of fine-tuning, which allows the classifier to learn based on the representations extracted by the pre-trained model. After this initial period, we unfreeze the encoder and aggregator and fine-tune them alongside the classifier.
Following prior works \cite{deldari2022cocoa,jain2022collossl,haresamudram2021contrastive}, we use the macro F1-score (unweighted mean of F1-scores over all classes) as the evaluation metric, as suggested for imbalanced datasets \cite{plotz2021applying}.

\subsection{Baselines}
To compare the effectiveness and capability in learning informative representations with no or limited labelled data, we investigate the performance of other baselines in different setups:

\noindent
\textbf{Fully supervised}: We evaluate the performance of \METHOD{} against two fully supervised baselines: DeepConvLSTM \cite{ordonez2016deep} and ``Supervised''. DeepConvLSTM is a widely used model architecture in HAR, and ``Supervised'' is the supervised equivalent of \METHOD~, in that it uses the same underlying architecture as \METHOD~ but is trained in a fully supervised manner. Both supervised baselines were trained using an end-to-end approach. For the DeepConvLSTM method, we utilize the implementation from a pre-existing source \cite{hoelzemann2020digging}.

\noindent
\textbf{Self-supervised baselines}: To evaluate the performance of \METHOD \space against other state-of-the-art self-supervised learning (SSL) models, we use COCOA \cite{deldari2021time} as a SOTA cross-modal SSL model. Other SSL baselines proposed for wearable sensor data, as discussed in Section \ref{sec:related}, were not considered due to their lack of cross-modality or inferior performance when compared to COCOA, which has been shown to outperform numerous baselines. 

\noindent
\textbf{Fixed and fine-tuned SSL encoders}: We examine the performance of the representations in two different setups: (1) fixed and (2) fine-tuned encoders, following the evaluation procedure described in \cite{henaff2020data, deldari2022cocoa}. During pre-training, all encoders were trained in a self-supervised manner. In the Fixed setup, during the classifier training step, these encoders are frozen in order to evaluate the quality of the learned representations; while in the fine-tuned setup, the encoders are re-trained (fine-tuned) along with the classifier based on the downstream task. 
We evaluate the effectiveness of the learned representations using a linear classifier and a Softmax layer, following the evaluation framework outlined in \cite{tian2019contrastive}. 

\begin{table}[htbp]
  \centering
  \caption{Performance comparison across three different setups: fully supervised, fixed SSL and fine-tuned SSL encoders.
  }
  \label{tab:baseline_results}
  \color{\revisioncolor}
  \resizebox{0.48\textwidth}{!}{%
    \begin{tabular}{l|l|c|c|c}
\multicolumn{2}{c|}{Technique}                                                          & \multicolumn{3}{c}{Dataset}                                       \\ \hline
\multicolumn{1}{l|}{Training}                                 & Method                  & SleepEDF             & PAMAP2               & WESAD               \\ \midrule
\multicolumn{1}{l|}{\multirow{2}{*}{End-2-End}}               & Supervised              & 0.717 (.03)          & 0.879 (.12)          & 0.884 (.02)         \\
\multicolumn{1}{l|}{}                                         & DeepConvLSTM            & 0.601 (.02)          & 0.718 (.18)          & 0.791 (.04)         \\ \midrule
\multicolumn{1}{l|}{\multirow{3}{*}{SSL (Fixed encoder)}}      & COCOA                   & 0.628 (.02)          & 0.839 (.11)          & 0.669 (.01)         \\
\multicolumn{1}{l|}{}                                         & \METHOD{} (random mask)  & 0.628 (.00)          & 0.802 (.15)          & 0.642 (.02)         \\
\multicolumn{1}{l|}{}                                         & \METHOD{} (spatial mask) & 0.722 (.02)          & 0.822 (.13)          & 0.667 (.02)         \\ \midrule
\multicolumn{1}{l|}{\multirow{3}{*}{SSL (Fine-tuned encoder)}} & COCOA                   & 0.678 (.01)          & 0.882 (.11)          & 0.913(.03) \\
\multicolumn{1}{l|}{}                                         & \METHOD{} (random mask)  & 0.726 (.00)          & 0.871 (.11)          & 0.894 (.02)         \\
\multicolumn{1}{l|}{}                                         & \METHOD{} (spatial mask) & \textbf{0.741 (.00)} & \textbf{0.892 (.10)} & \textbf{0.939 (.03)}       \\
\bottomrule
\end{tabular}%
}
\end{table}

\subsection{Comparison at One Glance}
Table \ref{tab:baseline_results} presents the average F1-score 
accompanied by their standard deviations. The optimal batch size (ranges from $8$ to $64$) for each baseline is reported. The results demonstrate that fine-tuning \METHOD~ surpasses the current state-of-the-art SSL methods and outperforms fully supervised baselines. 

\revision{\textbf{{\em Spatial vs. Random masking.}}} In terms of the masking strategy, spatial masking is clearly beneficial compared to the random masking strategy \revision{ by providing $2.06$\%, $2.04$\%, and $5.3\%$ higher f1-score in Fine-tuned \METHOD{} and $15.0\%$, $2.5\%$, and $5.3\%$ in Fixed \METHOD{} across SLEEPEDF, PAMAP2, and WESAD datasets, respectively.} As we mentioned, the intuition behind spatial masking is that one or more sources of data are not available hence the whole data for those sensors will be masked. For example, some users may not have some of the sensing devices or the device is temporarily switched off. While in random masking, the sensors are available but not all the time; hence the data can be sparse. For example, data is not available or useful at a particular moment due to energy issues or noise. We hypothesize that the inferior performance of random masking in comparison to spatial masking may be attributed to the model's need for continuity in the input data to derive meaningful information. Due to the random nature of the masking process, there may be several short segments of unmasked data, which may not provide sufficient information for the model to effectively learn.

\revision{\textbf{{\em Fixed vs. Fine-tuned SSL.}}} In the case of Fixed encoders, \METHOD\ with {\em random} masking strategy provides competitive results with COCOA but still cannot catch up with supervised baselines. On the other hand, once \METHOD~ is trained with spatial masking, it fairly improves the f1-score and can outperform the supervised baseline in the SLEEPEDF dataset. Once we fine-tune the self-supervised pre-trained COCOA, as well as \METHOD(random masking) and \METHOD~(spatial masking), they provided much higher performance compared to their corresponding Fixed setup. Based on Table \ref{tab:baseline_results}, this fine-tuning of COCOA, \METHOD(random), and \METHOD~(spatial) yields a increase in F1-score up to $7.9$\%, $15.5$\%, and $2.6$\% in SLEEPEDF dataset, $5.12$\%, $8.6$\%, and $8.5$\% over PAMAP2 dataset, and by $36.5\%$, $39.2\%$, and $38.9\%$ over WESAD dataset, respectively.

\revision{Overall, the fine-tuned \METHOD~ with spatial masking outperforms the fully supervised baseline by $3.3$\% and $1.5$\%, and $4.6\%$ across SLEEPEDF, PAMAP2, and WESAD datasets, respectively. Moreover, \METHOD{} improves the state-of-the-art cross-modal SSL model, COCOA, by $9.3$\% and $11.2$\%, $1.3\%$ across SLEEPEDF, PAMAP2, and WESAD datasets, respectively. WESAD shows the lowest delta in performance which can be attributed to the number of modalities as it includes only three sensors. On the other hand, the highest performance gain is with the PAMAP2 dataset which includes the highest number of modalities (seven sensors). This result further validates the superiority of \METHOD{} in
 datasets with a higher number of modalities and hence a higher chance of missing data.}

\subsection{Robustness Against Missing Data}
Even though missing data is one of the most critical challenges in wearable and ubiquitous computing, it is less studied in the existing literature. 
In this section, we evaluate the impact of missing data at fine-tuning and inference times. The main goal is to assess the robustness of \METHOD~ against the permanent or temporal absence of data sources. Based on the evaluation outcome, we can decide the best approach in training downstream tasks where missing data is unavoidable. The result from the other baseline COCOA is not included in this section due to the incapability of COCOA and other SOTA multimodal SSL models to handle missing data. 
To investigate this, we design three sets of experiments:
 {\bf (1) No missing.} The training and test sets are perfectly well-curated, with no missing data. We provide this as the upper-bound performance.
{\bf (2) Missing data only at inference time:}  In these experiments, we assume the model has access to well-curated data with all data points and sensors available. However, given there is less control over capturing the data at inference time, some devices or sensors may be absent (switched off or not present).
{\bf (3) Missing data at training and inference time:} We assume both training and test sets may contain missing data.
\revision{To make a fair comparison, we pre-train the encoders using both {\em random} and {\em spatial} masking strategies. We apply the pretrained model to the downstream task assessing different missing data scenarios.}
Table \ref{tab:missing_sensors_experiments} reports the average f1-score across five repeated runs with randomly missing data.

\begin{table}[htbp]
  \centering
  \caption{Evaluating across various missing modality scenarios}
  \label{tab:missing_sensors_experiments}
  \color{\revisioncolor}
  \resizebox{0.48\textwidth}{!}{%
\begin{tabular}{cccl|ccc}
\multicolumn{4}{c|}{Technique}                                                                                                             & \multicolumn{3}{c}{Dataset}                                                           \\ \hline
\multicolumn{2}{c}{Missing data at:}        & \multicolumn{1}{c|}{\multirow{2}{*}{masking}} & \multicolumn{1}{c|}{\multirow{2}{*}{Method}} & \multirow{2}{*}{SleepEDF} & \multirow{2}{*}{PAMAP2} & \multirow{2}{*}{WESAD}          \\ \cline{1-2}
Fine-tuning          & Inference            & \multicolumn{1}{c|}{}                         & \multicolumn{1}{c|}{}                        &                           &                         &                                 \\ \midrule
\multirow{5}{*}{No}  & \multirow{5}{*}{No}  & \multicolumn{1}{c|}{-}                        & Supervised                                   & 0.717 (.03)               & 0.879 (.12)             & 0.884 (.02)                     \\ 
                     &                      & \multicolumn{1}{c|}{\multirow{2}{*}{random}}  & Fixed SSL                                    & 0.628 (.00)               & 0.709 (.18)             & 0.629 (.02)                     \\
                     &                      & \multicolumn{1}{c|}{}                         & Fine-tuned SSL                               & 0.726 (.00)               & 0.825 (.13)             & 0.890 (.01)                     \\
                     &                      & \multicolumn{1}{c|}{\multirow{2}{*}{spatial}} & Fixed SSL                                    & 0.722 (.02)               & 0.822 (.14)             & 0.715 (.06)                     \\
                     &                      & \multicolumn{1}{c|}{}                         & Fine-tuned SSL                               & \textbf{0.741 (.00)}      & \textbf{0.892 (.11)} & \textbf{0.925 (.03)}\\ \midrule
\multirow{5}{*}{No}  & \multirow{5}{*}{Yes} & \multicolumn{1}{c|}{-}                        & Supervised                                   & 0.703 (.03)               & 0.897 (.11)             & 0.894 (.02)                     \\
                     &                      & \multicolumn{1}{c|}{\multirow{2}{*}{random}}  & Fixed SSL                                    & 0.602 (.03)               & 0.742 (.18)             & 0.622 (.03)                     \\
                     &                      & \multicolumn{1}{c|}{}                         & Fine-tuned SSL                               & 0.738 (.03)               & 0.859 (.13)             & 0.899 (.02)                     \\
                     &                      & \multicolumn{1}{c|}{\multirow{2}{*}{spatial}} & Fixed SSL                                    & 0.694 (.01)               & 0.805 (.16)             & 0.655 (.02)                     \\
                     &                      & \multicolumn{1}{c|}{}                         & Fine-tuned SSL                               & \underline{0.739 (.02)}   & \textbf{0.899 (.09)}    & \textbf{0.923 (.03)}\\ \midrule
\multirow{5}{*}{Yes} & \multirow{5}{*}{Yes} & \multicolumn{1}{c|}{-}                        & Supervised                                   & 0.202 (.17)               & 0.469 (.36)             & 0.304 (.37)                     \\
                     &                      & \multicolumn{1}{c|}{\multirow{2}{*}{random}}  & Fixed SSL                                    & 0.206 (.35)               & 0.331 (.19)             & 0.186 (.16)                     \\
                     &                      & \multicolumn{1}{c|}{}                         & Fine-tuned SSL                               & 0                         & 0.440 (.28)             & 0.139 (.18)                     \\
                     &                      & \multicolumn{1}{c|}{\multirow{2}{*}{spatial}} & Fixed SSL                                    & \underline{0.667 (.13)}   & \underline{0.646 (.21)} & \underline{0.278 (.14)}         \\
                     &                      & \multicolumn{1}{c|}{}                         & Fine-tuned SSL                               & 0.581 (.24)               & 0.495 (.35)             & 0.234 (.17)                     \\     
\bottomrule
\end{tabular}%
}
\end{table}

\begin{figure*}
\centering
\fcolorbox{white}{white}{
    \subfigure[][]{
    \label{fig:sleep_mask_random}
     \includegraphics[width=0.26\linewidth]{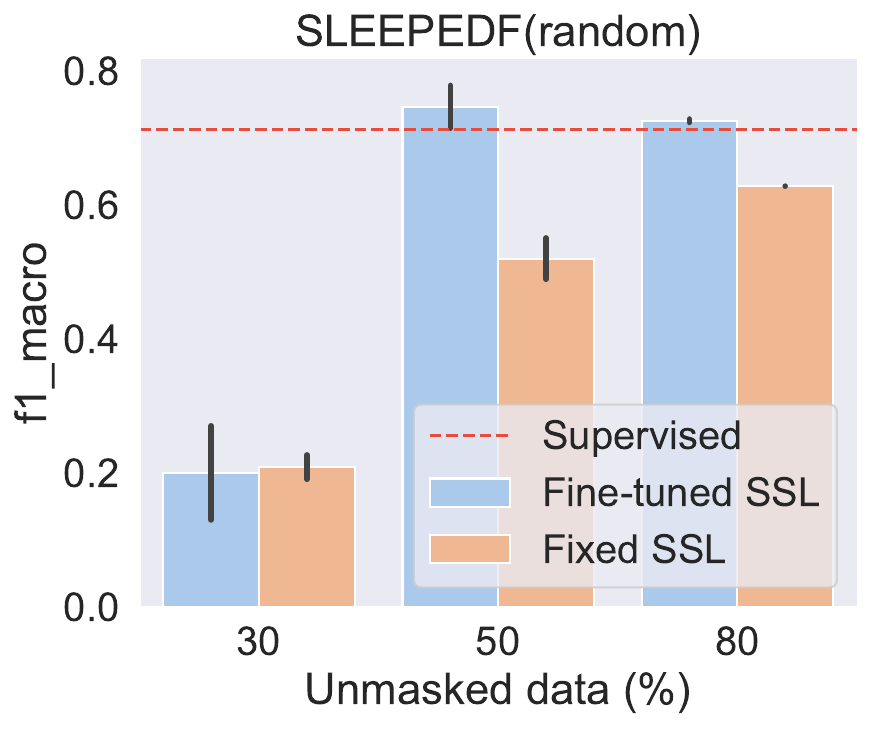}}
     \subfigure[][]{
    \label{fig:pamap_mask_random}
     \includegraphics[width=0.26\linewidth]{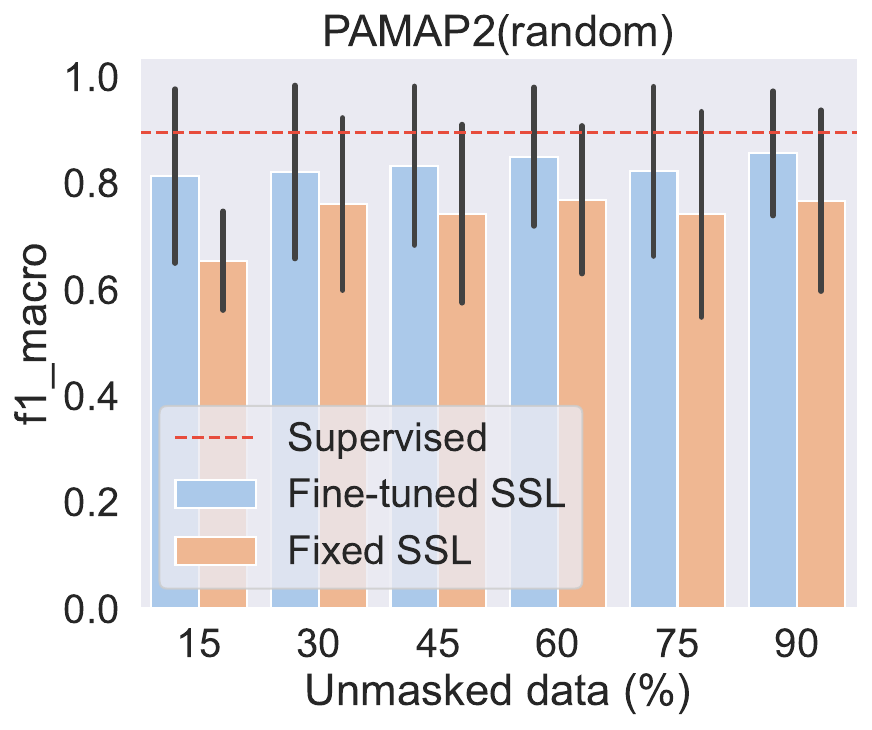}}
     \subfigure[][]{
    \label{fig:wesad_mask_random}
     \includegraphics[width=0.26\linewidth]{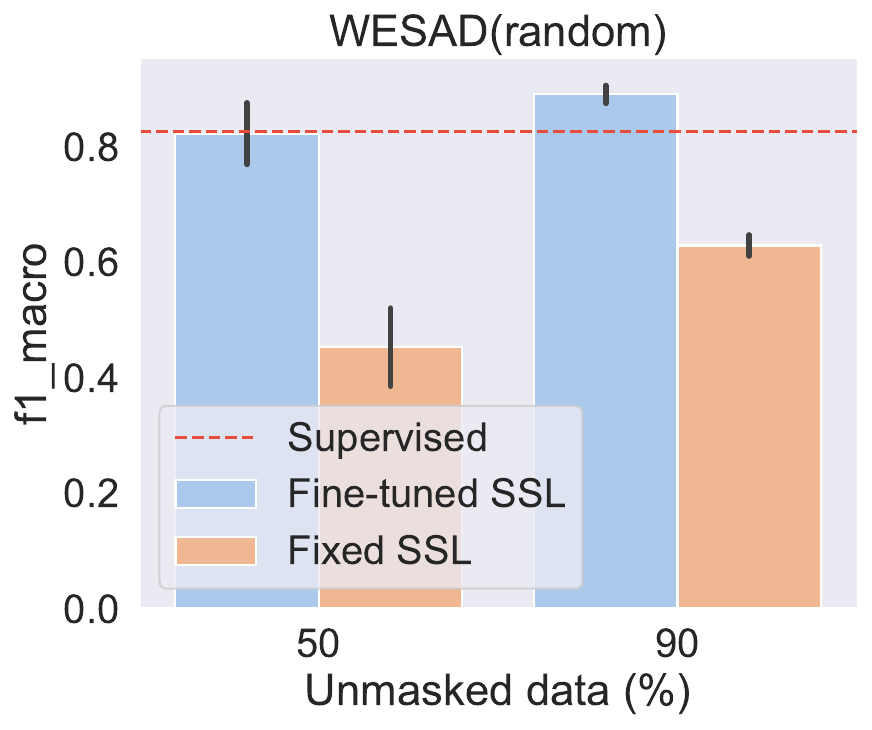}}}
\fcolorbox{white}{white}{     
     \subfigure[][]{
     \label{fig:sleep_mask_spatial}
     \includegraphics[width=0.26\linewidth]{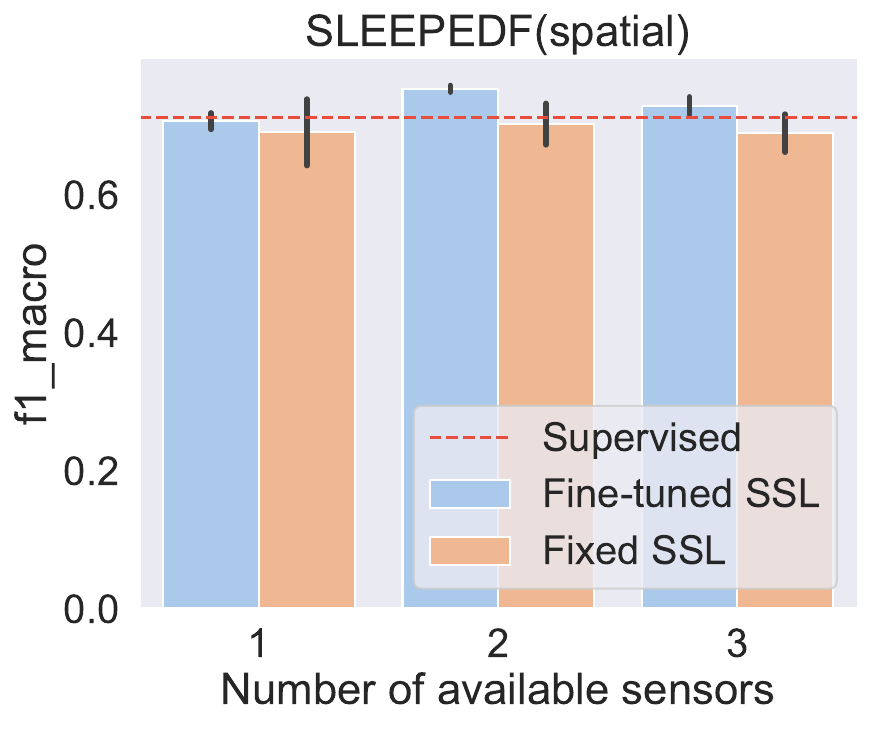}}
     \subfigure[][]{
     \label{fig:pamap_mask_spatial}
     \includegraphics[width=0.26\linewidth]{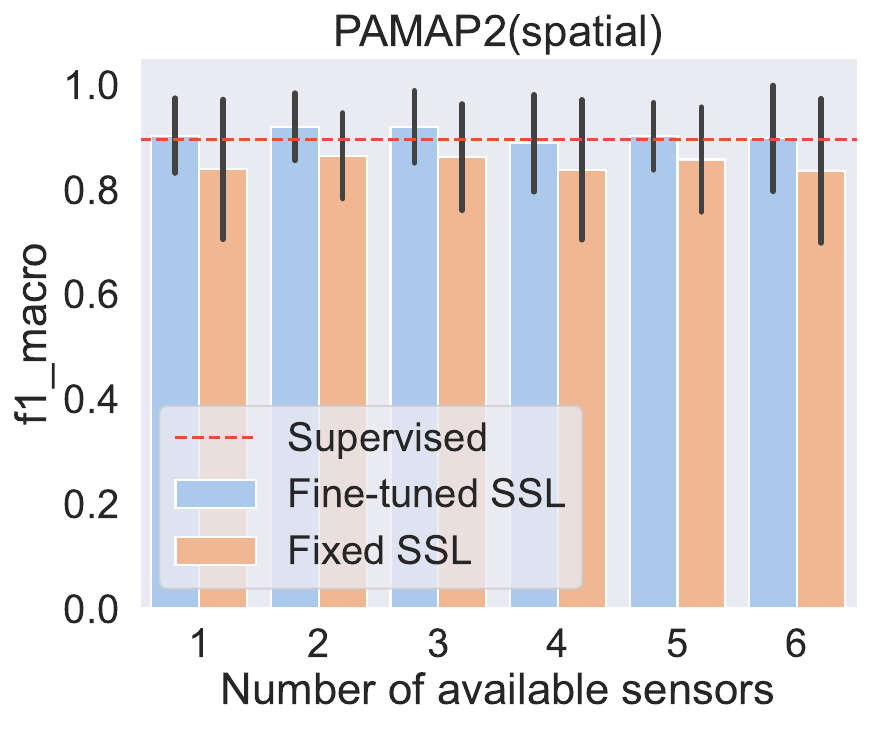}}
     \subfigure[][]{
     \label{fig:wesad_mask_spatial}
     \includegraphics[width=0.26\linewidth]{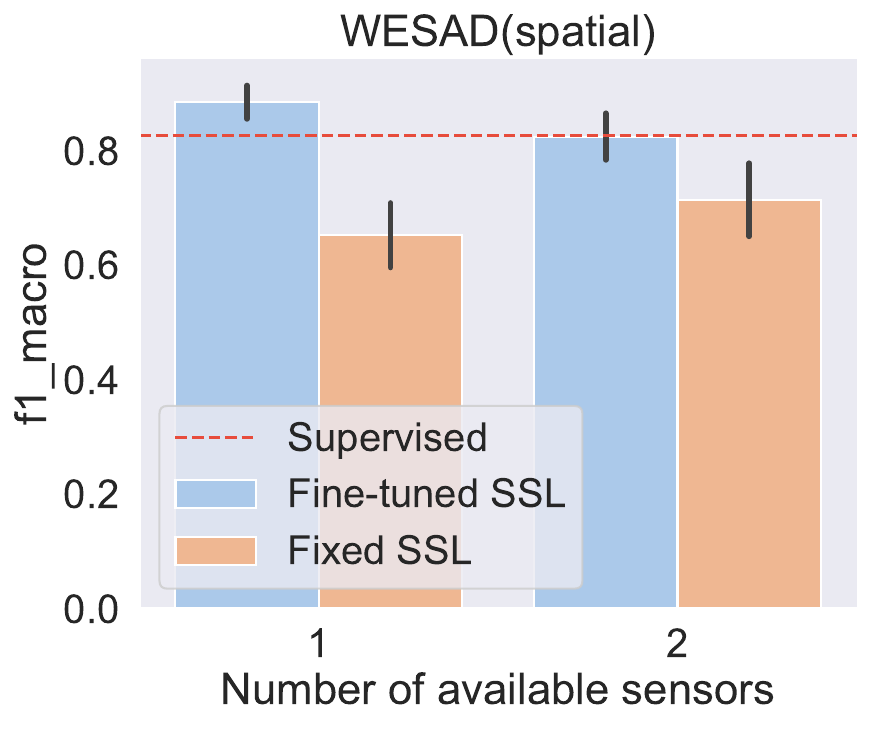}}
}
    \caption{Comparing random (a,b,c) and spatial (d,e,f) latent masking across the three datasets for different masking rates.}
    \label{fig:masking_all}
\end{figure*}

\revision{As shown in Table \ref{tab:missing_sensors_experiments}, Fine-tuned \METHOD~ with {\em spatial} masking pre-training and no missing data at the fine-tuning stage achieves the best performance compared to the other setups and datasets, including the fully-supervised model. In case of missing data at inference, the fine-tuned \METHOD~ can reach the same level of performance on both PAMAP2 and WESAD and a small drop on SleepEDF. The results validate its robustness to missing data at inference time. On the other hand, once the model is trained with all available data, it can handle missing data much better at inference time, compared to the previous scenario which introduced instability with the (un)availability of sensors at the fine-tuning training set.}

In the case of missing data at both fine-tuning and inference stage, we observe a considerable drop in the performance compared to the fully supervised model: $0.51$, $0.41$, and $0.58$ decrease in F1-score on SleepEDF, PAMAP2, and WESAD, respectively. Although the lower performance is inevitable, focusing on the fine-tuned and fixed variants of \METHOD with {\em spatial} masking, \METHOD{} is particularly effective. The performance of \METHOD with Fixed encoders is not affected much by missing data at fine-tuning step and it outperforms both supervised and fine-tuned models, highlighting the role of a well-trained encoder for high-quality representations.

\begin{figure*}
\centering
\fcolorbox{white}{white}{
     \subfigure[][]{
     \label{fig:sleep_label_spatial}
     \includegraphics[width=0.28\linewidth]{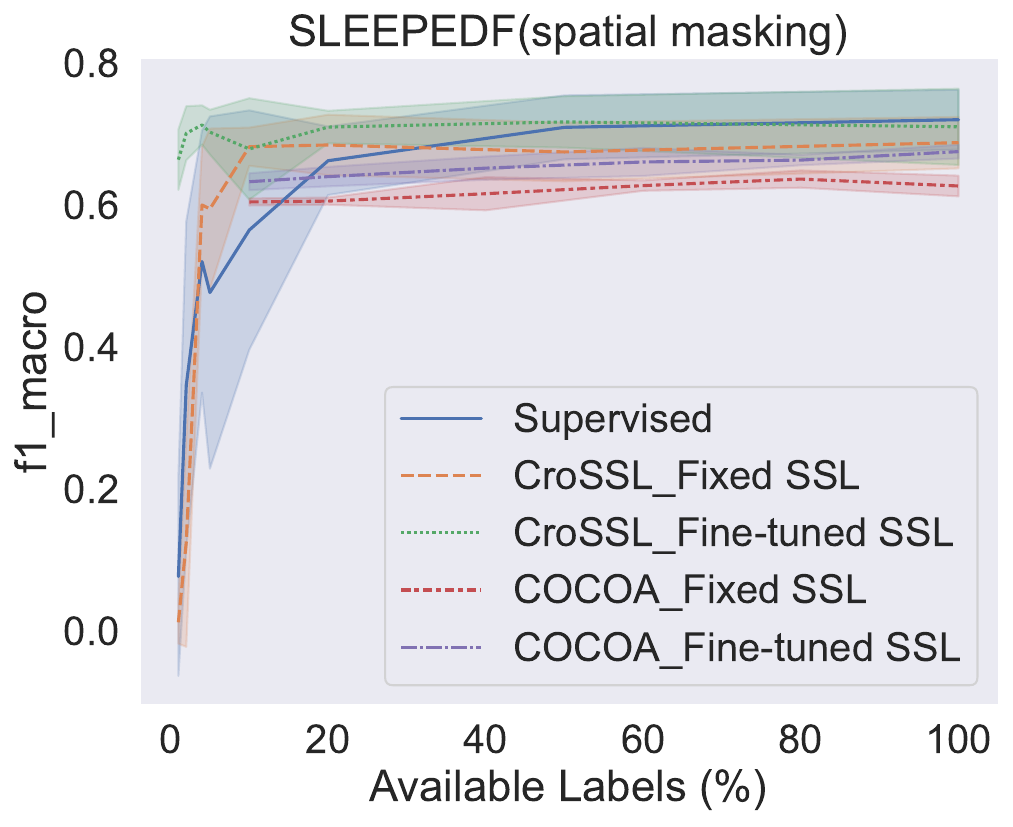}}
     \subfigure[][]{
     \label{fig:pamap_label_spatial}
     \includegraphics[width=0.28\linewidth]{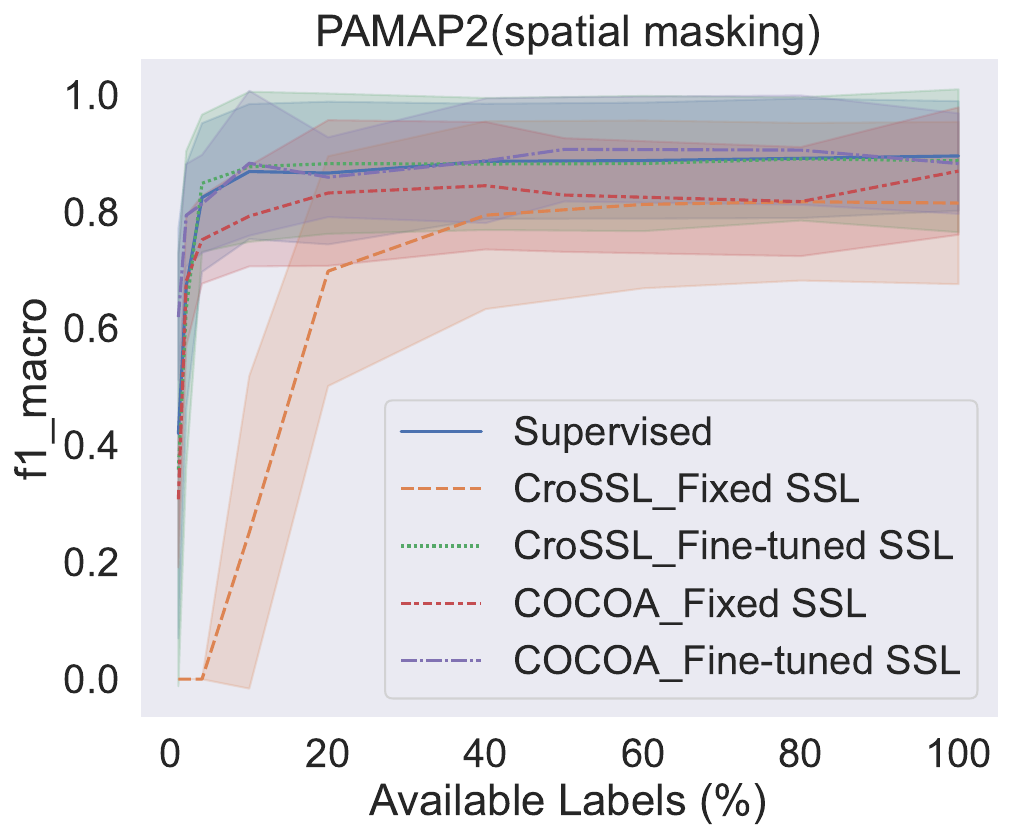}}
     \subfigure[][]{
     \label{fig:pamap_label_spatial}
     \includegraphics[width=0.28\linewidth]{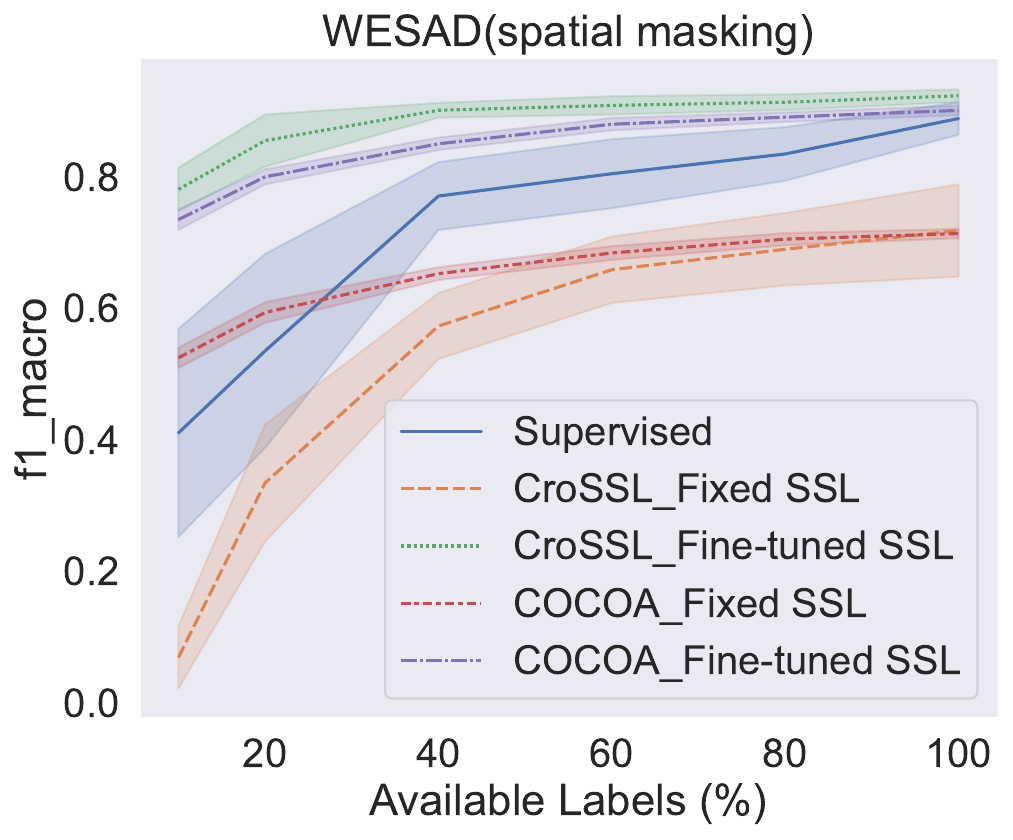}}}
    \caption{Comparing the efficiency of fixed and fine-tuned \METHOD~ setups against fully supervised model in low-labeled data}
    \label{fig:label_efficiency}
\end{figure*}

\subsection{Optimal Masking Strategies} 
\label{sec:masking_section}
As latent masking is central to \METHOD, we investigate the impact of missing data ratio. As an alternative to entirely random masking, we also apply spatial masking in order to assess whether hiding entire modalities increases the robustness of the model. Figure \ref{fig:masking_all} shows fine-tuned models outperform fixed ones across all fronts.

\revision{\textbf{\textit{Random Masking.}} We observe a slight negative correlation between the masking rate and the quality of learnt representations across all datasets. Due to the lack of any fine-tuning in Fixed SSL, the power of the pre-trained encoders in learning useful representations comes into action. However, in the PAMAP2 dataset, this correlation is weaker, which can be attributed to the type of sensors and the distribution of data. Given that PAMAP2 contains data from 3 accelerometers, 3 gyroscopes, and a heart rate sensor, there is more shared information among the input channels. Thus, even a subset of sensors can still retain the essential information about the activity. This finding further confirms our motivations in the theory section above (\ref{sec:theory}). }
An inverse trend is observed in SleepEDF, where higher masking corresponds to lower performance. Notably, the highest performance is achieved with a 50\% masking rate which hints at a U-shaped relationship for the fine-tuned variant. We do not observe such trends in the fixed variant.

\revision{\textbf{\textit{Spatial Masking.}}} As discussed earlier, spatial masking achieves the best performance in our experiments. We do not observe significant differences for most datasets when varying the number of spatial modalities. \revision{ In particular, the mean performance peaks with 2, 3, and 1 sensors in SLEEPEDF, PAMAP2, and WESAD datasets, respectively (however, the std overlaps do not allow for conclusive results). WESAD presents the largest differences between 1 and 2 sensors, with a decrease in performance after applying Fine-tuning but an increase after applying a Fixed encoder. In all datasets, there is a slight decrease in performance with the most sensors available.}

These results validate the value of the latent masking, especially for bigger datasets. We note that the SleepEDF dataset is five times larger than PAMAP2 and we could attribute its performance results --with regards to masking-- to this size difference. In other words, SSL requires enough samples for pre-training, and in light of these results, we hypothesize that masking makes this task even harder by hiding latent information. Therefore, we expect the effect of masking to be stronger with larger pre-training datasets.

\subsection{Label-efficiency}
\label{sec:label_efficiency_sec}
We investigate the efficacy of \METHOD~ in low-labelled data scenarios compared to its self-supervised and fully supervised counterparts. 
Figure \ref{fig:label_efficiency} presents the average F1-score of \METHOD~ with fixed pretrained and also fine-tuned encoders along with the fully supervised baseline \revision{ where the size of labelled data varies between $1$\% and $100$\% of the available labels for SleepEDF and PAMAP2, and $10\%$ and $100\%$ for the WESAD dataset}. We investigated the impact of the masking strategy and reported the best one (spatial masking in all cases). For this experiment, we train the classifier over self-supervised encoders by using only a subset of labelled data. 

Fixed \METHOD~ is extremely label-efficient. Using spatial masking, Fixed \METHOD~  achieves its optimal performance via less than half of labels in each dataset (\ie $10\%$, $20$\%, and $60$\% of the available labels in the SleepEDF, PAMAP, and WESAD).  This showcases the capability of our pre-trained model, which can be combined with a simple MLP classifier to attain peak performance with limited labeled data. This feature is encouraging for the deployment on edge devices that have limited resources and the ability to train and maintain large machine-learning models. \revision {Compared to COCOA, \METHOD{} with fixed encoder setup shows lower performance in lower labeled data regimes. We believe this happens due to the less available data for training in \METHOD{} due to the applied masking.}

Fine-tuned \METHOD~ (with both spatial and random masking) shows great improvement compared to the fully supervised model in a low-labeled data regime, specifically for SleepEDF and WESAD. Given only $1$\% of labels, the fine-tuned model with spatial masking almost achieves its highest F1-score while the supervised model achieves the same using at least $20$\%  and $40\%$ of labels in SleepEDF and WESAD, respectively.
The PAMAP2 exhibits a higher standard deviation compared to the other datasets due to the variation among users in the respective test sets and its smaller size. This is because, for each evaluation fold, we only included one user per test, resulting in higher variation. In contrast, the SleepEDF dataset has multiple users dedicated to each test set, leading to lower variation.

\section{Discussion and Limitations}
\METHOD \space puts forward a general ML framework for sensor time series that achieves SOTA performance in multimodal benchmark tasks spanning from activity recognition to sleep stage classification. 
Notably, \METHOD \space is data and label-efficient, requiring only a small fraction of labeled samples to achieve performance on par or better than supervised models. Most importantly, our model proposed the latent masking idea that ensures that the model learns robust representations in an end-to-end manner, without requiring any data pre-processing. We showed that latent masking is effective and makes models more accurate and transferable. 

In the current implementation of \METHOD{}, we used two masking strategies: spatial and random. Our results suggest that spatial masking is preferable. Future work could investigate temporal or spatiotemporal masking, however, it is not straightforward to design such masks due to the loss of sequence information in the latent space. Further, the design of custom masking strategies could leverage domain knowledge of the signals, such as existing interactions between the modalities \cite{spathis2021self}.  
Another exciting property of SSL methods is transferring across various tasks. 
\cite{haresamudram2022assessing} have evaluated within the SSL paradigm focusing on a single modality while changing the sensor positions, activities, and sampling rates \cite{haresamudram2022assessing}.

\section{Conclusion}
Given the pervasiveness of sensor-enabled devices, there are increasing interest in designing applications that leverage multiple modalities of data. An important first step toward this is to design learning algorithms that can extract high-quality embeddings from multimodal data, even in the absence of labels. To this end, we presented \METHOD{}, a novel self-supervised learning technique to train global embeddings from multimodal sensor streams by leveraging the spatiotemporal correlations in them. To address the challenge of heterogeneous ubiquitous sensor computing applications, \METHOD{} employs sensor-specific encoders with the possibility of taking different sample sizes. This makes the model invariant to the type of input modalities and is able to capture sensor-specific information.
Our key findings are that \METHOD{} outperforms fully-supervised and state-of-the-art self-supervised approaches on \revision{three} challenging datasets while remaining robust to missing modalities. Using a masking-based technique, \METHOD{} forces the model to learn representations invariant to the presence of all modalities.
Moreover, \METHOD{} is highly label-efficient and hence can be deployed in applications where data labeling is expensive.

\begin{acks}
The authors affiliated with UNSW would like to acknowledge the support of Cisco’s National Industry Innovation Network (NIIN) Research Chair Program.
\end{acks}

\bibliographystyle{ACM-Reference-Format}
\bibliography{ref}

\end{document}